\documentclass[conference]{IEEEtran}

\usepackage{cite}
\usepackage{graphicx}
\usepackage{xcolor}
\usepackage{subfigure}
\usepackage[hidelinks]{hyperref}
\usepackage{booktabs}
\usepackage{multirow}
\usepackage{siunitx} 

\newcommand{\ie}{\textit{i.e.}}
    
\begin{document}

\title{Scalable Smartphone Cluster for Deep Learning
}

\author{
\IEEEauthorblockN{Byunggook Na$^{1}$, Jaehee Jang$^{1}$, Seongsik Park$^{2}$, Seijoon Kim$^{3}$, Joonoo Kim$^{4}$,\\
Moon Sik Jeong$^{4}$, Kwang Choon Kim$^{4}$, Seon Heo$^{4}$, Yoonsang Kim$^{4}$, Sungroh Yoon$^{1,5,\ast}$}
\IEEEauthorblockA{
$^{1}$Dept. of Electrical and Computer Engineering, Seoul National University, Seoul, South Korea \\
$^{2}$Korea Institute of Science and Technology, Seoul, South Korea \\
$^{3}$Samsung Advanced Institute of Technology, Samsung Electronics, Suwon, South Korea \\
$^{4}$Mobile Communications Business, Samsung Electronics, Suwon, South Korea \\
$^{5}$ASRI, INMC, ISRC, and Interdisciplinary Program in AI, Seoul National University, Seoul, South Korea \\
$^{\ast}$sryoon@snu.ac.kr 
}
}

\maketitle

\begin{abstract}

Various deep learning applications on smartphones have been rapidly rising, but training deep neural networks (DNNs) has too large computational burden to be executed on a single smartphone.
A portable cluster, which connects smartphones with a wireless network and supports parallel computation using them, can be a potential approach to resolve the issue.
However, by our findings, the limitations of wireless communication restrict the cluster size to up to 30 smartphones.
Such small-scale clusters have insufficient computational power to train DNNs from scratch.
In this paper, we propose a scalable smartphone cluster enabling deep learning training by removing the portability to increase its computational efficiency.
The cluster connects 138 Galaxy S10+ devices with a wired network using Ethernet.
We implemented large-batch synchronous training of DNNs based on Caffe, a deep learning library.
The smartphone cluster yielded 90\% of the speed of a P100 when training ResNet-50, and approximately 43x speed-up of a V100 when training MobileNet-v1.

\end{abstract}


\section{Introduction}

\begin{figure*}[!t]
    \centering
    \subfigure[]{
    \includegraphics[width=0.64\linewidth]{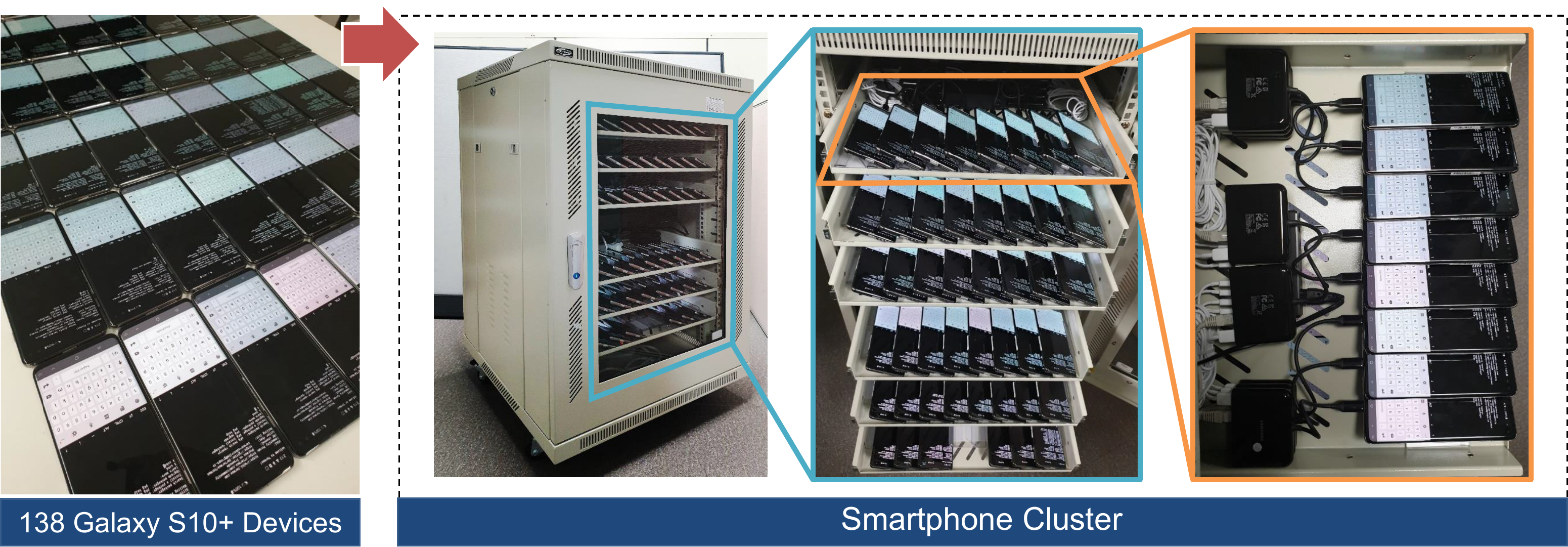}
    \label{subfig:cluster}
    }
    \subfigure[]{
    \includegraphics[width=0.16\linewidth]{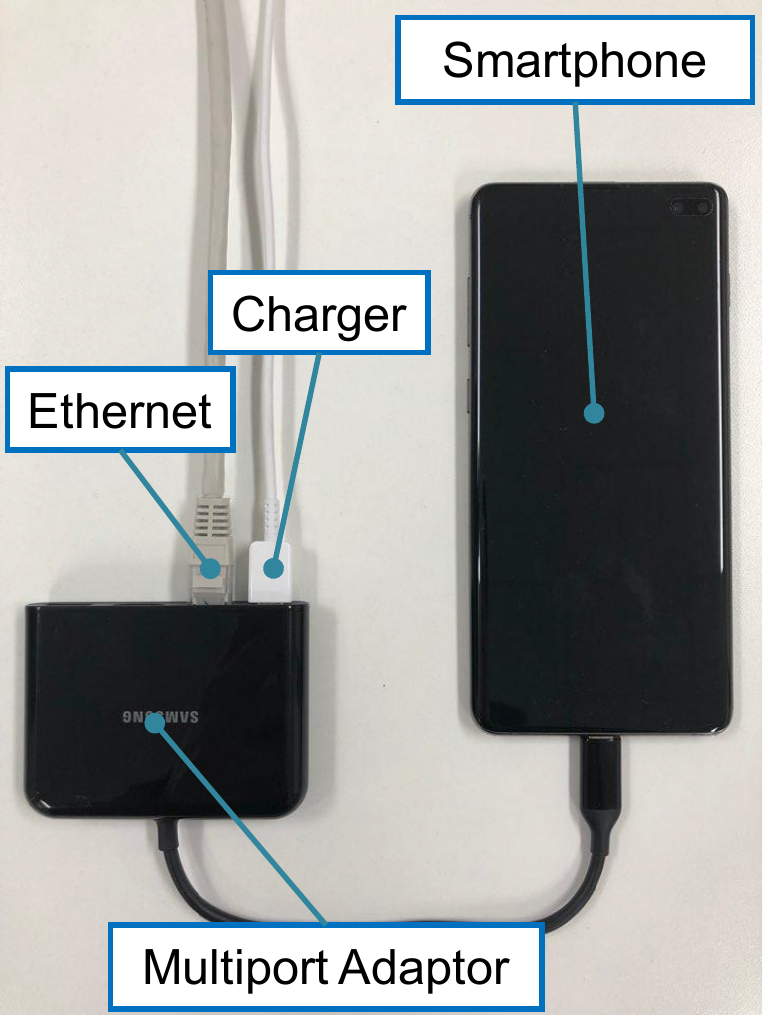}
    \label{subfig:configuration}
    }
    \subfigure[]{
    \includegraphics[width=0.125\linewidth]{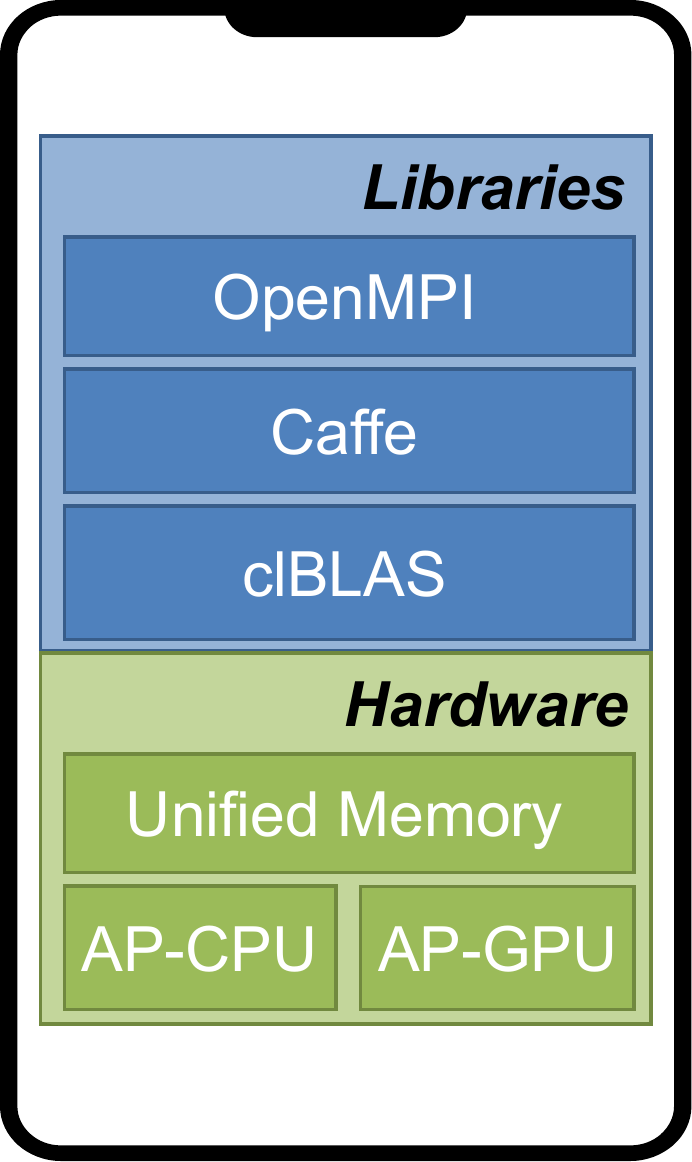}
    \label{subfig:node_stack}
    }
    \vspace{-1em}
    \caption{The proposed smartphone cluster for deep learning. (a) A single rack of 46 Galaxy S10+ devices is presented, and we constructed three racks using 138 devices. (b) A multi-port adaptor used in the cluster provides power and transfers data to the corresponding smartphone. (c) Hardware and software stacks of our distributed deep learning implementation on smartphones is revealed.}
    \vspace{-1em}
\end{figure*}

Various deep learning applications have run on mobile devices such as smartphones~\cite{Xu2019AFL}.
Deep learning has been mostly studied on general-purpose graphic processing units (GPGPUs) because of its large amount of computational power.
Smartphones have much less powerful processors and less memory than GPGPUs.
Nevertheless, it is possible to perform deep learning on smartphones by reducing neural network sizes and computations~\cite{lane2016deepx, hubara2017quantized, park2018quantized, lin2017runtime, he2017channel,iandola2016squeezenet,howard2017mobilenets,tan2019mnasnet,han2016eie,park2019energy}.
However, existing approaches have largely focused on the inference phase of deep learning.
There are still insufficient studies on ways to train deep neural networks (DNNs) on smartphones, which require much more computations than inference.
By adopting transfer learning, finetuning pre-trained DNNs on a single smartphone can be executed with smaller amount of data and computation~\cite{valery2018low} than training the DNNs from scratch.

Portable clusters of mobile devices have been proposed to be used for parallel computation~\cite{busching2012droidcluster,arslan2015cwc, kumar2016powershare, attia2016high, mao2017modnn, mao2017adalearner}. 
However, the cluster construction techniques are usually based on the assumption that the mobile devices should remain portable.
Smartphones of these portable clusters are typically connected by wireless networks, limiting computational power of the clusters to run distributed algorithms because of the small scalability and low data transmission rate of the wireless networks.
Furthermore, only small-scale clusters consisting of less than ten devices have been demonstrated, and these clusters have insufficient computational power to train DNNs that are usually trained on GPGPUs.

Larger clusters connected in a way that promotes computational effectiveness have the potential to train DNNs at speeds comparable to GPGPUs, if portability is not a priority.
To construct the larger clusters, we can use redundant smartphones that do not belong to anyone.
Smartphones include harmful chemicals when released to the environment. Furthermore, the amount of smartphone waste is huge since replacement period for a smartphone is generally one or two years.
The obsolete smartphones can cause serious threat to the environment.
Therefore, as a motivation, our re-purposing approach of the smartphones can highly contribute to eco-friendliness, which is one of the most important and fundamental topics for the future.

We constructed a smartphone cluster which can accommodate up to 138 Galaxy S10+ devices.
We believe that this is the first practical scheme to use more than tens of smartphones for parallel computation.
To overcome the limited connectivity of wireless networks, we employed a wired network using Ethernet and connected the smartphones with multi-port adaptors, which support the data transfer and a supply of power.
On this cluster, we implemented a distributed deep learning program using the OpenMPI, a message passing interface (MPI) library, and Caffe, a deep learning library supporting OpenCL.

We evaluated computational efficiency obtained with different numbers of smartphones with a fixed mini-batch size, and identified that a strategy based on the data-parallelism of traditional distributed deep learning algorithms~\cite{dean2012large} cannot fully utilize the computation power of the smartphones in our cluster.
We therefore suggest using large-batch training methods and choosing mini-batch sizes which fully utilize the device memory.
Compared with recent GPGPUs, when training ResNet-50, our cluster achieved 91\%, 66\%, and 55\% of the speed of P100, V100, and 2080ti GPGPUs, respectively.
Remarkably, it outperformed these GPGPUs with approximately 35x, 43x, and 22x speed-up when training MobileNet-v1.
By expanding the cluster at large scale (e.g., thousands of smartphones), we anticipate that the computational performance gap is reduced and it can even make our cluster faster than GPGPUs.

\section{Related Work}
Several methods have been proposed to assemble a portable smartphone cluster anywhere.
DroidCluster~\cite{busching2012droidcluster} demonstrated the feasibility of a portable cluster with only six mobile devices by evaluating it using LINPACK benchmark for high performance computing systems.
Kumar et al.~\cite{kumar2016powershare} built a cluster of four mobile devices and evaluated it computationally simple tasks such as counting numbers.
Attia et al.~\cite{attia2016high} performed a matrix multiplication on two mobile devices.
This is much simpler than deep learning training, which requires many matrix multiplications.

The use of portable clusters for deep learning has been demonstrated at small-scale~\cite{mao2017modnn,mao2017adalearner,mao2017mednn}.
Mao and his colleagues~\cite{mao2017modnn, mao2017mednn} performed deep learning inference rather than training on four smartphones, and also~\cite{mao2017adalearner} showed that basic DNNs such as LeNet was able to be trained on eight smartphones.
These authors did not suggest that they could train non-trivial DNNs capable of more accurate classification performance.

Larger clusters may have a potential to train the DNNs.
We have been able to locate only one attempt~\cite{arslan2015cwc} to connect more than ten mobile devices.
This study~\cite{arslan2015cwc}, which involved 18 smartphones that were dispersed into different places, was focused on algorithms to manage the smartphones in a way that accommodates fluctuating connectivity, during the execution of a distributed computation.

In the aforementioned studies, the authors connected their devices using wireless networks such as WiFi or Bluetooth.
However, wireless networks are slow and the connectivity is likely to be intermitted when many smartphones are connected.
This makes it hard to expand the cluster and perform tasks that are too heavy to run on portable clusters, such as distributed deep learning training which involves large amounts of computations as well as a lot of data transmission.

\section{Method}

The requirements to be met by a scalable smartphone cluster for deep learning training are: 
(1) high data transmission rate connections;
(2) an appropriate infrastructure for hardware stability of the power supply and temperature; 
(3) a distributed deep learning framework with low communication requirements for the scalability.
We designed a smartphone cluster to satisfy these requirements, and constructed it from 138 Galaxy S10+ smartphones, as depicted in Fig.~\ref{subfig:cluster}.

\subsection{Connections among Smartphones}
\begin{figure}
    \centering
    \includegraphics[width=\linewidth]{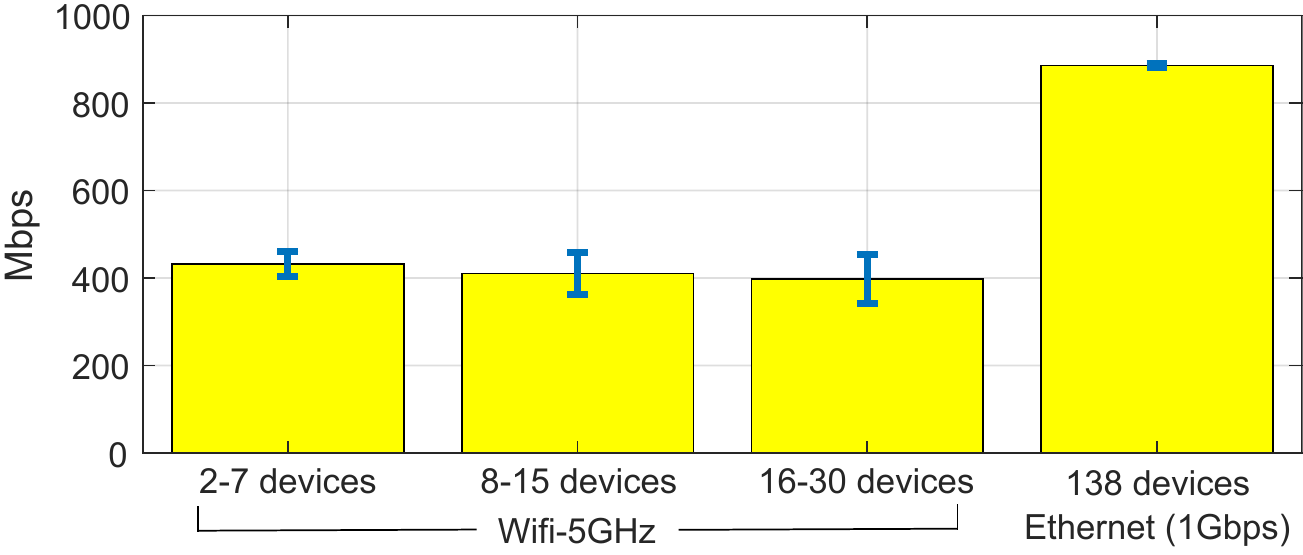}
    \vspace{-6.0mm}
    \caption{One-to-one communication speed of wireless and wired networks. Average and standard deviation of 10 runs, measured with \textsf{iperf3}. When more than 30 smartphones were connected using WiFi-5GHz, these connected smartphones were usually disconnected enough to abandon the tests. In contrast, when using Ethernet, 138 smartphones are stably connected and fast data transmission between the devices is supported.}
    \label{fig:iperf3}
\end{figure}
\begin{figure}
    \centering
    \includegraphics[width=\linewidth]{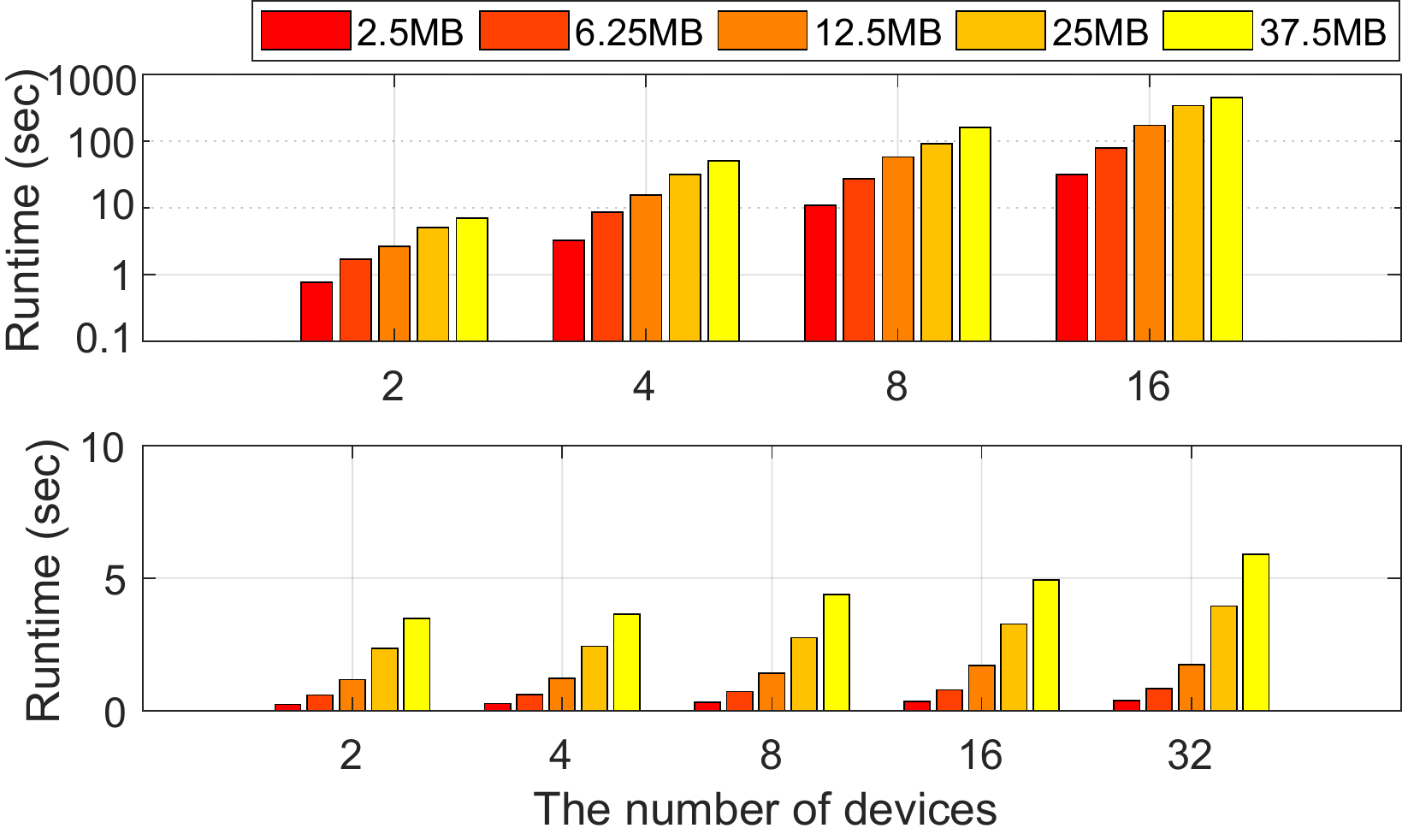}
    \vspace{-6.0mm}
    \caption{Communication time of \texttt{MPI\_allreduce} of OpenMPI running on wireless and wired networks. Results of WiFi-5GHz are much slower and less scalable than those of Ethernet.}
    \label{fig:mpi_allreduce}
    \vspace{-3.0mm}
\end{figure}

Smartphones are capable of both wireless and wired communications. 
To enable distributed deep learning, high speed and stable connection are crucial because of the high communication demand and long training time of the distributed deep learning training. 
We started by looking at the performance of wireless connections within a cluster incorporating more device than those of previous authors~\cite{busching2012droidcluster, arslan2015cwc, kumar2016powershare, attia2016high, mao2017modnn, mao2017adalearner}.
We compared the one-to-one communication speed of WiFi-5GHz and Gigabit Ethernet by measuring the data transmission rate using \textsf{iperf3} within \textsf{Termux}, which is a linux emulator running on Android OS; the results are presented in Fig.~\ref{fig:iperf3}.

When smartphones are connected by WiFi-5GHz, connecting more smartphones reduced communication speed (lower average) and made it less predictable (higher standard deviation). 
Attempts to connect more than 30 smartphones resulted in high rates of disconnection, and thus we had to inevitably abandon the tests.
We concluded that WiFi-5GHz is not a suitable connection environment for a large smartphone cluster.
Conversely, we successfully expanded our smartphone cluster size to 138 smartphones through three 48-port Ethernet switch hubs, even with no reduction in one-to-one communication speed.
The Ethernet connections were more than two times faster and more stable than WiFi-5GHz.

We conducted additional experiments using the smartphones and collective primitives of OpenMPI: one-to-many and many-to-many communication tests with various quantities of data, using functions such as \texttt{MPI\_reduce} and \texttt{MPI\_allreduce}.
The results of \texttt{MPI\_allreduce} shown in Fig.~\ref{fig:mpi_allreduce} reveals again that WiFi-5GHz is much slower and less scalable than Ethernet.
As the number of smartphones changed from two to 16 to execute \texttt{MPI\_allreduce} on 37.5MB data, communication time of WiFi-5GHz and Ethernet become 63 times and 1.3 times longer, respectively.
We therefore build our smartphone cluster using Ethernet to connect 138 smartphones.

\subsection{Stable Environment for Deep Learning}
For the stable training, constant heat management and power supply is fundamental. 
Training DNNs is a computationally intensive task which consumes large amount of power, and thus elevates the temperature of smartphones that generally lack cooling systems used in servers equipped with GPGPUs.
Such a temperature rise of a smartphone can reduce the overall computational performance of the cluster due to the thermal throttling~\cite{bhat2019power}.
Thermal throttling is a technique to protect processors and users from heat damage by reducing the clock frequency of processors inside the smartphones but resulting in the drop of the computational power.
In addition, the large power consumption of the deep learning can turn off the smartphones, resulting in that the deep learning training is discontinued.
To support the stable environment for deep learning, appropriate system design is required.

We tested two setups for constant power supply without redundant temperature rise:
1) using a wireless charger and a separate Ethernet adaptor,
and 2) using a multi-port adaptor for both power and Ethernet.
The use of setup 1 caused the temperature of a fully charged device to increase by \SI{5}{\celsius}, where the device is even idle. 
This may advance the triggering of thermal throttling.
Additionally, we found the wireless chargers unreliable; the smartphones easily became disconnected with the wireless chargers by small movements of the smartphone cluster.
We therefore adopted setup 2 using multi-port adaptors, as shown in Fig.~\ref{subfig:configuration}.
This allowed us to place the smartphones at an angle, which improved airflow being able to be augmented with fans and thus reduced thermal throttling.

\subsection{Deep Learning on a Smartphone Cluster}
\begin{figure}[t]
    \centering
    \includegraphics[width=\linewidth]{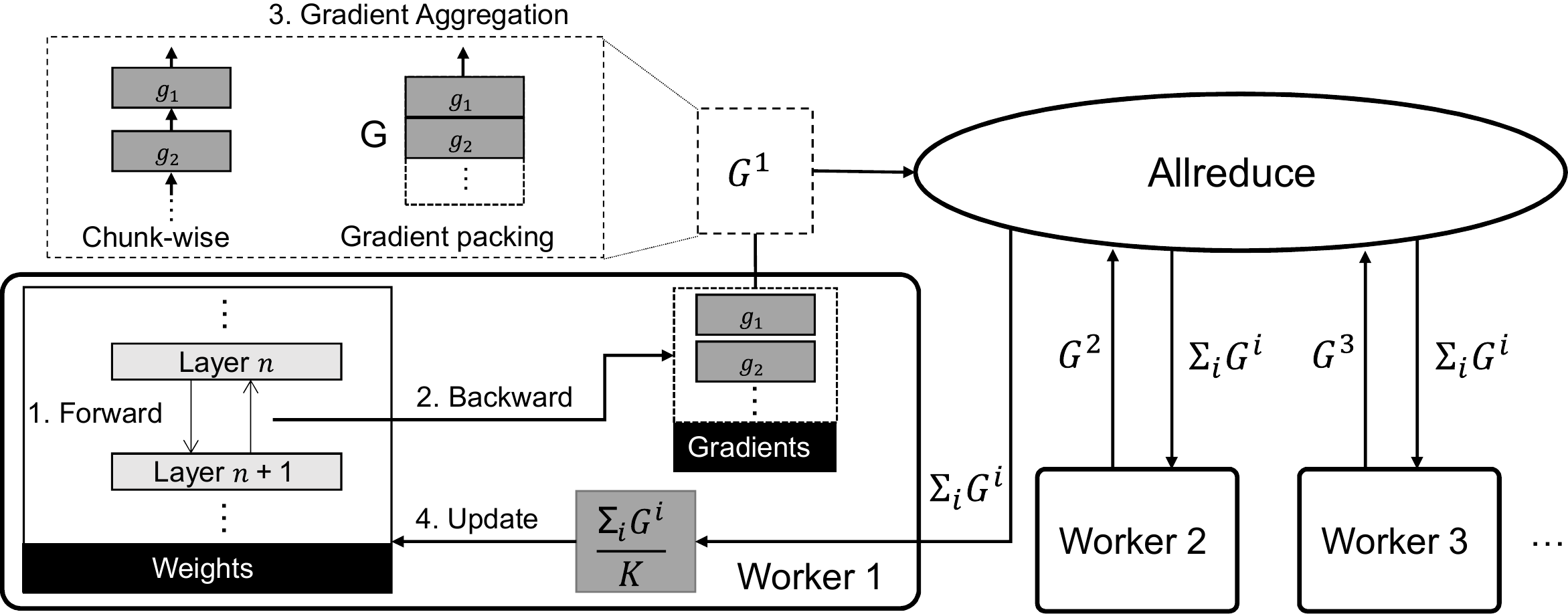}
    \vspace{-3.0mm}
    \label{fig:flowchart}
    \caption{Flowchart of a distributed deep learning using the allreduce operation for gradient aggregation when $K$ smartphones are used. The forward and backward calculations and the update process are based on Caffe, and the gradient aggregation is implemented with OpenMPI.}
    \vspace{-3.0mm}
\end{figure}
%
Synchronous training~\cite{dean2012large} is a fundamental distributed deep learning technique, based on data parallelism; the training data is divided into workers (\ie, smartphones in our context) that synchronously train their local models.
As shown in Fig.~\ref{fig:flowchart}, the synchronous training has two phases: computation and communication.
In the computation phase, the workers execute forward and backward calculations to obtain gradients used in the deep learning training using local models and local data
Gradients that are calculated by every worker are aggregated and then transmitted to all workers during the communication phase.
Workers update the weights of their local models using the aggregated gradients.

Our deep learning framework for smartphones, based on synchronous training, is an extension of the widely used deep learning framework, Caffe, which is written in C/C++ and supports OpenCL.
Efficient communication during the gradient aggregation phase is provided by data transfer functions from the OpenMPI library.
Gradient aggregation is performed by a modified implementation of the ring-based allreduce (RAR) operation~\cite{patarasuk2009bandwidth}.
The smartphones in the cluster have the OpenCL library to support the multicore computing capabilites of the CPUs and a GPU found inside AP chips, denoted by AP-CPUs and AP-GPU, respectively.
The OpenCL library supports several sets of basic linear algebra subroutines (BLAS), and we tested clBLAS, clBLAST and OpenBLAS for the computation.
From a comparison of runtimes, we decided that clBLAS was the best choice for Caffe and was executed as 4 threads on the AP-CPUs.
Our program was cross-complied on \textsf{Termux} for the AP chips.

We improved computational efficiency to adopt a technique referred as gradient packing.
As shown in Fig.~\ref{fig:flowchart}, our implementation of Caffe-based distributed deep learning algorithm transfers all the gradients as a single chunk of data to reduce the number of invocations of communication functions required.
Before the gradient aggregation step, the gradient of all the layers are copied into a chunk of memory of the collect size. 
Then, the chunks from each smartphone are transferred to the other smartphones by a single invocation of the RAR operation.
Such way of transferring gradients as a single chunk whose size is same to a DNN, is faster than a way how gradients of each layer are individually transferred, which is referred to as a chunk-wise way. 

\subsubsection{Remark}
We first chose Darknet~\footnote{http://pjreddie.com/darknet/} as our deep learning engine.
Darknet is also written C/C++ and supports OpenCL.
Contrary to Caffe, Darknet showed the best speed when it was executed with clBLAS on the AP-GPU.
We had struggles to improve computational efficiency of Darknet-based distributed deep learning.
For examples, using the zero-copy technique of OpenCL and taking advantage of unified memory in the APs, gradients calculated by the AP-GPU can be directly transferred to other smartphones without memory copy required in GPGPUs.
However, Darknet was less competitive than Caffe in terms of learning possibilities of various DNNs as well as processing speed on smartphones.

\section{Experimental Evaluation}
We constructed the smartphone cluster with 138 Galaxy S10+ development devices.
The Galaxy S10+ was equipped with the Qualcomm SM8150 AP, referred to as Snapdragon 855, and 6GB memory.
Note that 2.8GB were available due to memory requirements of Android OS and system applications. 
During the training, MPI processes were assigned to the smartphones in one-to-one way.

\subsection{Functionality Validation}
To validate functionality of our deep learning training implementation running on the smartphone cluster, we trained convolutional neural networks (CNNs) on CIFAR10~\cite{deng2009imagenet}.
Given the same neural network configurations and training hyperparamters, the training results were similar to those of GPGPUs.
For example, in the case of training GoogleNet on ImageNet for 96000 iterations, top-1 test accuracy results were respectively 59.6\% and 60.1\% on the smartphone cluster and Nvidia 2080ti, respectively; in this experiments, hyperparameters were 736 batch size, 0.01 learning rate, 0.0002 weight decay and step learning rate decay policy.

\subsection{Scaling Performance Analysis}
\begin{figure}
    \centering
    \includegraphics[width=\linewidth]{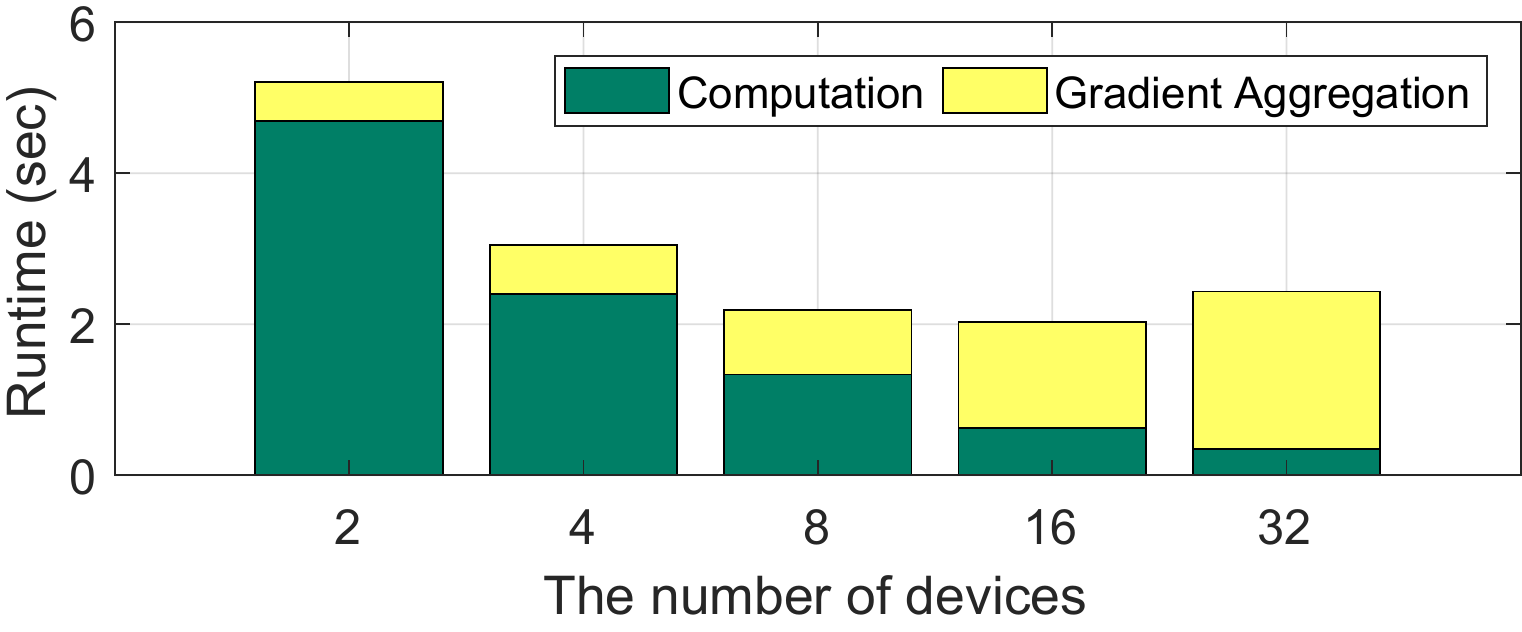}
    \vspace{-1em}
    \caption{Scaling experiments of Caffe-based distributed deep learning when the mini-batch size is fixed as 32 and GoogleNet model is used.}
    \label{fig:scaling}
    \vspace{-1em}
\end{figure}

We investigated scaling effect when a traditional strategy based on data-parallelism of distributed deep learning.
For the tests with GoogleNet, the mini-batch size was fixed as 32 and the number of smartphones participating the tests gradually increased by two times; that is, the local mini-batch size per worker decreased from 32 to one by two times.
Fig.~\ref{fig:scaling} reveals that the computation time decreased by two times as well, but conversely, communication time increased gradually, meaning that the scaling effect may plateau in terms of total runtime.
When using 32 smartphones, the total runtime is longer than when using 16 smartphones.
Therefore, such data-parallel strategy is not computational efficient, and limits to use more smartphones than the fixed mini-batch size; in this test, more than 32 smartphones cannot be used.

To relieve the limitation effectively, a strategy proposed by large-batch training~\cite{goyal2017accurate,you2018imagenet} is valid.
The large-batch training shortens training time by reducing the number of training iterations without training performance degradation.
If the training requires the fixed number of epochs, increasing mini-batch size reduces the number of training iterations.
Therefore, adopting the large-batch training techniques, we can set that the data size, denoted mini-batch-size-per-device, processed in a iteration on each device is maximized to fully use memory, resulting in that computational efficiency can be maximized.

\subsection{Efficiency Maximization Strategy}
\begin{figure*}
    \centering
    \includegraphics[width=\linewidth]{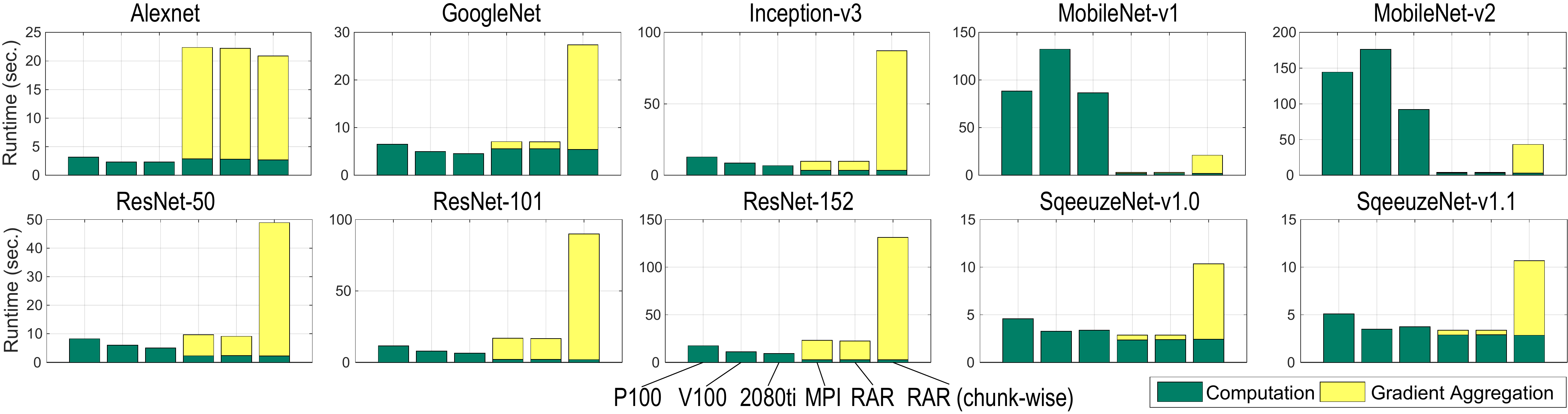}
    \caption{Evaluation for various models; for each model, we measured runtime per iteration processing data of the maximum mini-batch-size-per-device in Table~\ref{tab:model}. These figures are the results of three GPGPUs (P100, V100, 2080ti) and our smartphone cluster with three types of gradient aggregation (\texttt{MPI\_Allreduce} with gradient packing, ring-based allreduce with gradient packing, and ring-based allreduce with chunk-wise way).}
    \label{fig:caffe_time}
\end{figure*}
\begin{table}[t]
\caption{Evaluation DNNs}
\label{tab:model}
\centering
\vspace{-1em}
\begin{tabular}{crrr}
\toprule
\multirow{3}{*}{Model} & Size & Number of        & Mini-batch    \\
                       & (MB) & gradient chunks  & size per device \\
\midrule
AlexNet           & 232.56   &  16   & 32 \\
GoogleNet         & 26.70    &  116  & 16 \\
Inception-v3      & 91.05    &  556  & 4  \\
Mobilenet-v1      & 16.23    &  164  & 8  \\
Mobilenet-v2      & 13.51    &  320  & 8  \\
ResNet-50         & 97.70    &  321  & 4  \\
ResNet-101        & 170.34   &  626  & 2  \\
ResNet-152        & 230.20   &  932  & 2  \\
SequeezeNet-v1.0  & 4.76     &  52   & 16 \\
SequeezeNet-v1.1  & 4.71     &  52   & 32 \\
\bottomrule
\end{tabular}
\vspace{-2em}
\end{table}

To examine computational efficiency on various DNNs, we used models that are designed to train ImageNet and available at Caffe github.
Their detailed descriptions are listed up in Table~\ref{tab:model}, mini-batch-size-per-device of each DNN is set to maximize use of device memory.
Note that the sizes of all gradients, which are transmitted in distributed deep learning, are the same.
Experimental results obtained on 138 smartphones are in Fig.~\ref{fig:caffe_time}.

First, RAR gradient aggregation with a chunk-wise way is quite slower than RAR gradient aggregation with the gradient packing. 
Deeper DNNs tend to take more communication time of gradient aggregation when using the chunk-wise way.
For instance, while the sizes of Inception-v3 and ResNet-50 are similar, their communication time of RAR with the chunk-wise way is 84 and 47 seconds, respectively.
In contrast, for AlexNet that is relatively shallow, the chunk-wise way is slightly faster than the gradient packing. 
In practical, DNNs are generally deeper than AlexNet, thus gradient aggregation with gradient packing is more promising than that with the chunk-wise way.

We compared MPI\_Allreduce and RAR, in terms of communication time for the gradient aggregation. 
Both operations performed on 138 smartphones were very similar on most DNNs.
Meanwhile, when using 46 smartphones (its results are not visualized for the sake of the space), RAR was faster up to 1.56x for ResNet-152.

\begin{figure*}[ht]
    \centering
    \subfigure[]{
      \includegraphics[width=0.30\linewidth]{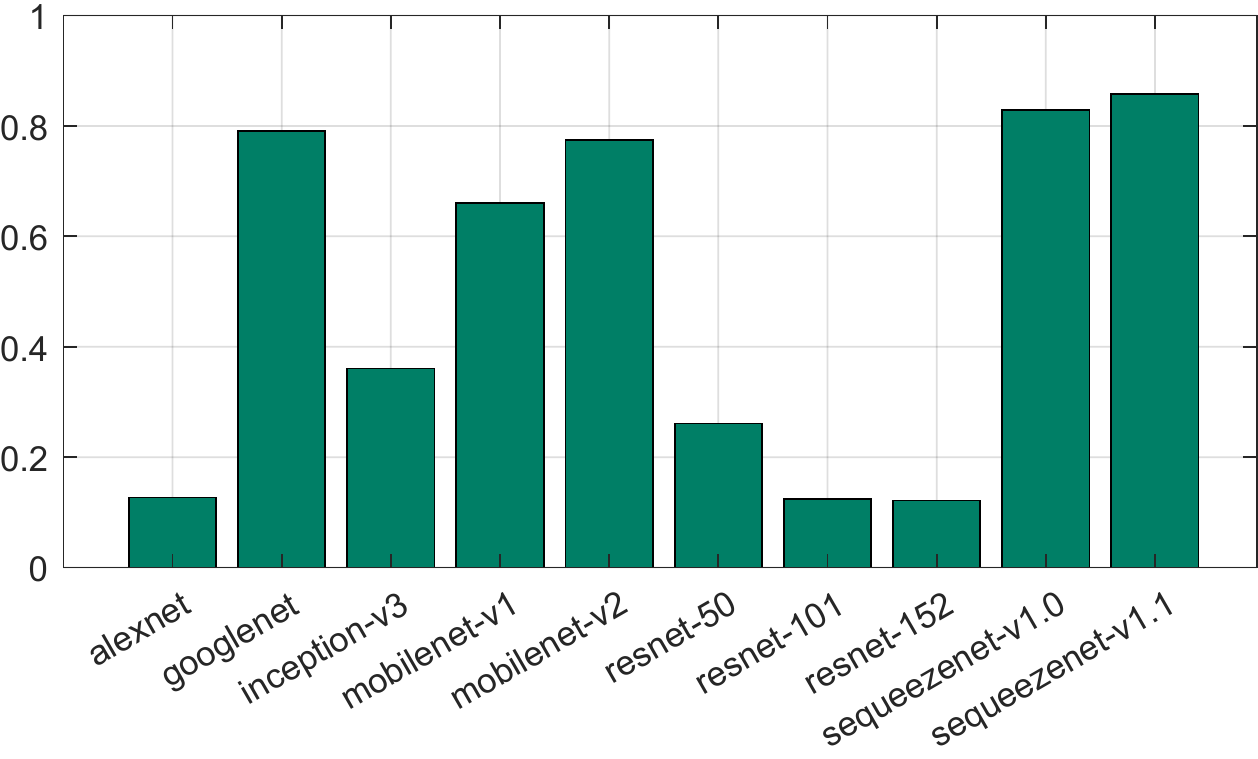}
      \label{fig:efficiency}
     }
    \subfigure[]{
      \includegraphics[width=0.27\linewidth]{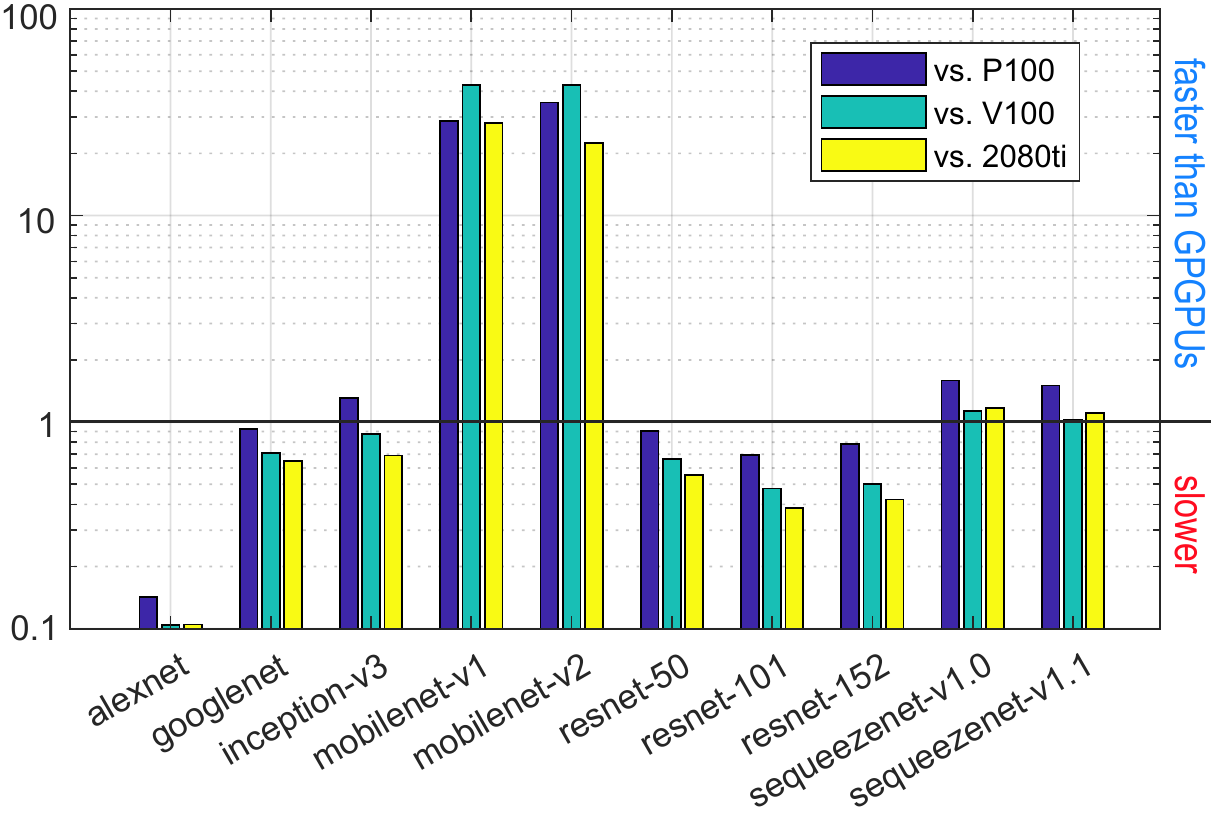}
      \label{fig:gpu_comparison}
    }
    \subfigure[]{
      \includegraphics[width=0.375\linewidth]{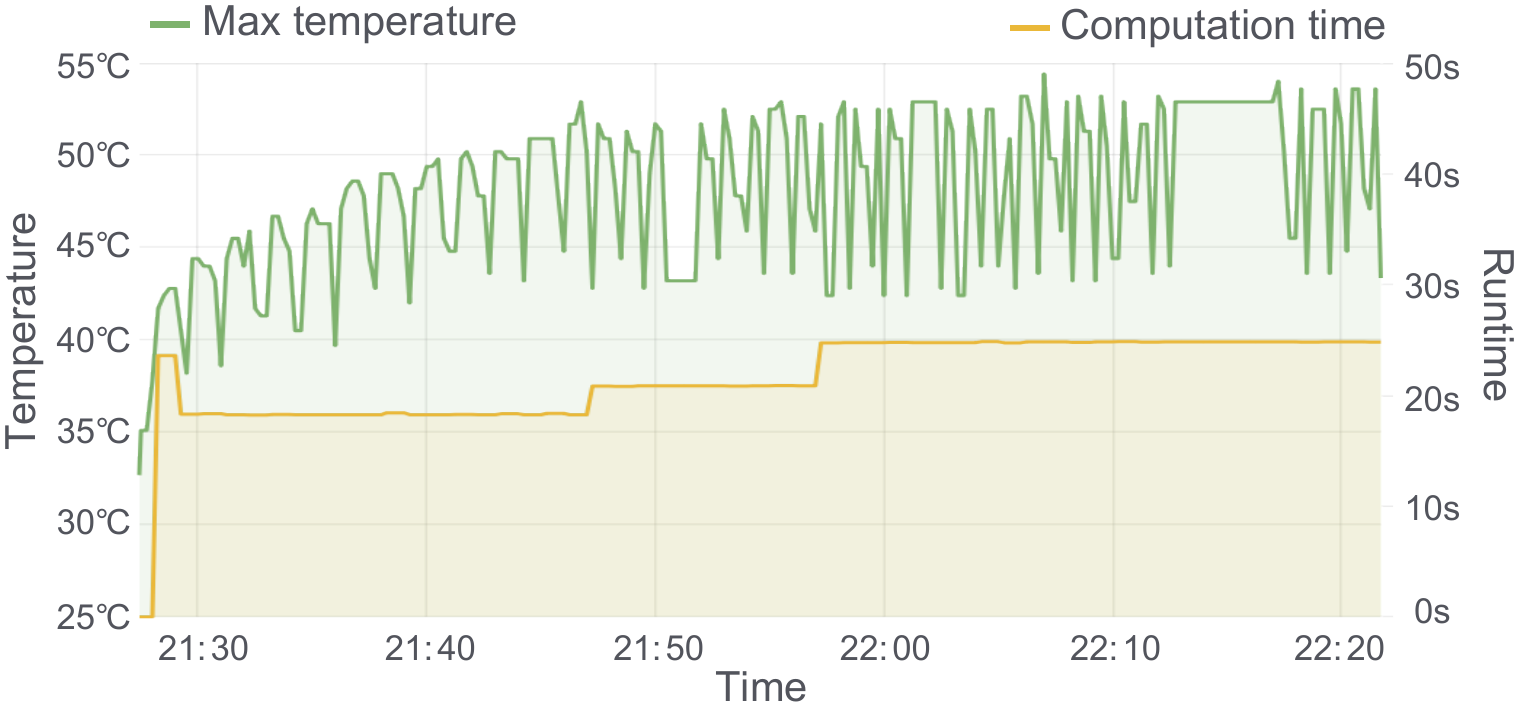}
      \label{fig:thermal_issue}
    }
    \vspace{-1em}
    \caption{(a) Computational efficiency of various DNNs on the smartphone cluster consisting of 138 smartphones, (b) relative processing speed of the smartphone cluster compared with GPGPUs (P100, V100, 2080ti), and (c) observed thermal throttling effect.}
    \vspace{-1em}
\end{figure*}

Fig.~\ref{fig:efficiency} presents computational efficiency of RAR cases in Fig.~\ref{fig:caffe_time} overall.
Evidently, as larger the model size is, the computational efficiency (the runtime of the computation within the total runtime) for the model is lower.
Our cluster achieved 85.8\% efficiency at most; SqueezeNet-v1.1.
The worst case was ResNet-152, which shows 12.2\% efficiency.
In the ResNet-152, a single device executed the computation phase of ResNet-152 for mini-batch-per-device-size of two due to memory limitation.
Thus, the computation time of such small data size is considerably shorter than transferring time of the gradients size (\ie, 230MB).

\subsection{Comparison with GPGPUs}
As a main motivation of our work, we investigated whether the smartphone cluster can serve as an alternatives to GPGPUs or not. 
For three recent GPGPUs (Nvidia P100, V100, and 2080ti), we measured processing time of the computation phase using the same mini-batch size (\ie, mini-batch-per-device-size $\times$ 138).
For the experiments conducted on GPGPUs, we used Caffe compiled with cuDNN.

The results are summarized as shown in Fig.~\ref{fig:gpu_comparison}; y-axis means how shorter processing time per iteration on the cluster than that on the GPGPUs.
Even though our cluster did not achieved to about 10\% of computation performance of GPGPUs on AlexNet, it showed notable strength on two versions of MobileNet~\cite{howard2017mobilenets,Sandler_2018_CVPR}; at most, 3525\%, 4298\% and 2244\% compared to the GPGPUs, respectively.
The MobileNet aims to reduce the size of the model to suit mobile devices. 
It includes a key component called depthwise separable convolution that was not optimized to utilize computational power of GPGPUs effectively.
As a result, although computational performance of the cluster was less than GPGPUs except some DNNs, this was the meaningful attempt to give a feasibility that the computational performance of the smartphone cluster can be comparable to that of GPGPUs, which has never been demonstrated in previous studies.

\section{Discussion}

\subsection{Toward Larger Scale}
We linked 138 smartphones into a cluster with a wired network and used it to train a DNN.
Our next goal is to construct a cluster consisting of thousands of smartphones which our organization can handle.
This is likely to involve addressing the following issues:
1. The increase in communication time.
2. compact racking of the smartphones while providing adequate cooling.
3. stable training without training performance degradation.

To maximize computational efficiency of the large cluster, it is critical to design network topologies suitable to such large scale as well as adequate collective primitives for maximizing throughput on the topologies.

\subsection{Thermal Impact on Computational Performance}
To access the cooling situation, we measured device temperatures and computation times during training without any forced cooling (Figure~\ref{fig:thermal_issue}).
As the temperature increased, computation phase of deep learning training took longer as the smartphones reduced their processor speeds by thermal throttling.
The computation time increased twice, 14.8\% and 36.3\% more than the initial computation time, 18.2 sec., respectively.

To minimize the thermal effect, while conducting the experiments reported in this paper, we used fans to cool down the cluster.
A more elaborate cooling system is required to sustain computational performance.

\subsection{Heterogeneous Mobile Device Cluster}
We used only Galaxy S10+ smartphones to construct our cluster.
But many different types of obsolete smartphones are available in practical, and a cluster could be made with smartphones with different amounts of computational power.
This would reduce the efficiency of synchronous training, because the runtime of gradient calculation (\ie, forward and backward processing in training) would be determined by the slowest smartphone.
Other approaches, such as asynchronous distributed deep learning~\cite{zhang2015deep}, might be expected to give better computational efficiency results in this case.

\section{Conclusion}
We constructed and demonstrated a scalable smartphone cluster by connecting 138 Galaxy S10+ smartphones with a wired network that is much faster, more stable and scalable than wireless networks.
To maximize the computational efficiency of our Caffe-based distributed deep learning, we employed large-batch training methods.
From the evaluation results on various DNNs, our smartphone cluster showed comparable performance to GPGPUs in training DNNs.
We plan to extend our approach to larger clusters, and modify distributed deep learning techniques to make them more suitable for smartphone clusters.
Our approach offers a significant contribution to sustainability by re-purposing the obsolete smartphones which would otherwise, be discarded.


\bibliography{references}
\bibliographystyle{IEEEtran}

\end{document}